\pdfoutput=1

\documentclass[11pt]{article}

\usepackage[]{ACL2023}

\usepackage{times}
\usepackage{latexsym}
\usepackage{booktabs}
\usepackage{subcaption}

\usepackage{amsmath}
\usepackage{amssymb}

\usepackage{xcolor}
\usepackage[T1]{fontenc}
\usepackage{hyperref}

\usepackage[utf8]{inputenc}

\usepackage{microtype}

\usepackage{inconsolata}

\usepackage{graphicx}
\usepackage{enumitem}
\usepackage{colortbl}
\usepackage{array,multirow}
\usepackage[symbol]{footmisc}

\title{Daisy-TTS: Simulating Wider Spectrum of Emotions\\via Prosody Embedding Decomposition}

\author{Rendi Chevi \\
  MBZUAI \\
  \texttt{rendi.chevi@mbzuai.ac.ae} \\ \And
  Alham Fikri Aji \\
  MBZUAI \\
  \texttt{alham.fikri@mbzuai.ac.ae} \\}

\begin{document}
\maketitle
\begin{abstract}
We often verbally express emotions in a multifaceted manner, they may vary in their intensities and may be expressed not just as a single but as a mixture of emotions. This wide spectrum of emotions is well-studied in the \textit{structural model of emotions}, which represents variety of emotions as derivative products of primary emotions with varying degrees of intensity. In this paper, we propose an emotional text-to-speech design to simulate a wider spectrum of emotions grounded on the structural model. Our proposed design, Daisy-TTS~\footnote[2]{We named the model ‘Daisy’ as a reference to the flower-like arrangement of the structural model of emotions. Interactive Demo: \url{https://rendchevi.github.io/daisy-tts/}}, incorporates a prosody encoder to learn emotionally-separable prosody embedding as a proxy for emotion. This emotion representation allows the model to simulate: (1) \textbf{Primary emotions}, as learned from the training samples, (2) \textbf{Secondary emotions}, as a mixture of primary emotions, (3) \textbf{Intensity-level}, by scaling the emotion embedding, and (4) \textbf{Emotions polarity}, by negating the emotion embedding. Through a series of perceptual evaluations, Daisy-TTS demonstrated overall higher emotional speech naturalness and emotion perceiveability compared to the baseline.
\end{abstract}


\section{Introduction}

Subtle variations in speaking patterns that make up our prosody, allow us to express a wide spectrum of emotions~\citep{cowen2019primacy}. In most cases, emotions may not always be expressed in their "pure" states~\citep{scherer2014approaches}. Emotions we express may vary in their intensity-level, for instance, a higher intensity of \textit{sadness} might be distinctly expressed as \textit{grief}. It may even get more intricate as sometimes we express not just a single but a mixture of different emotions~\citep{plutchik1982psychoevolutionary}.

\begin{figure}[!htb]
    \centering
    \includegraphics[trim={4.5cm 4cm 4cm 2cm},width=.9\linewidth]{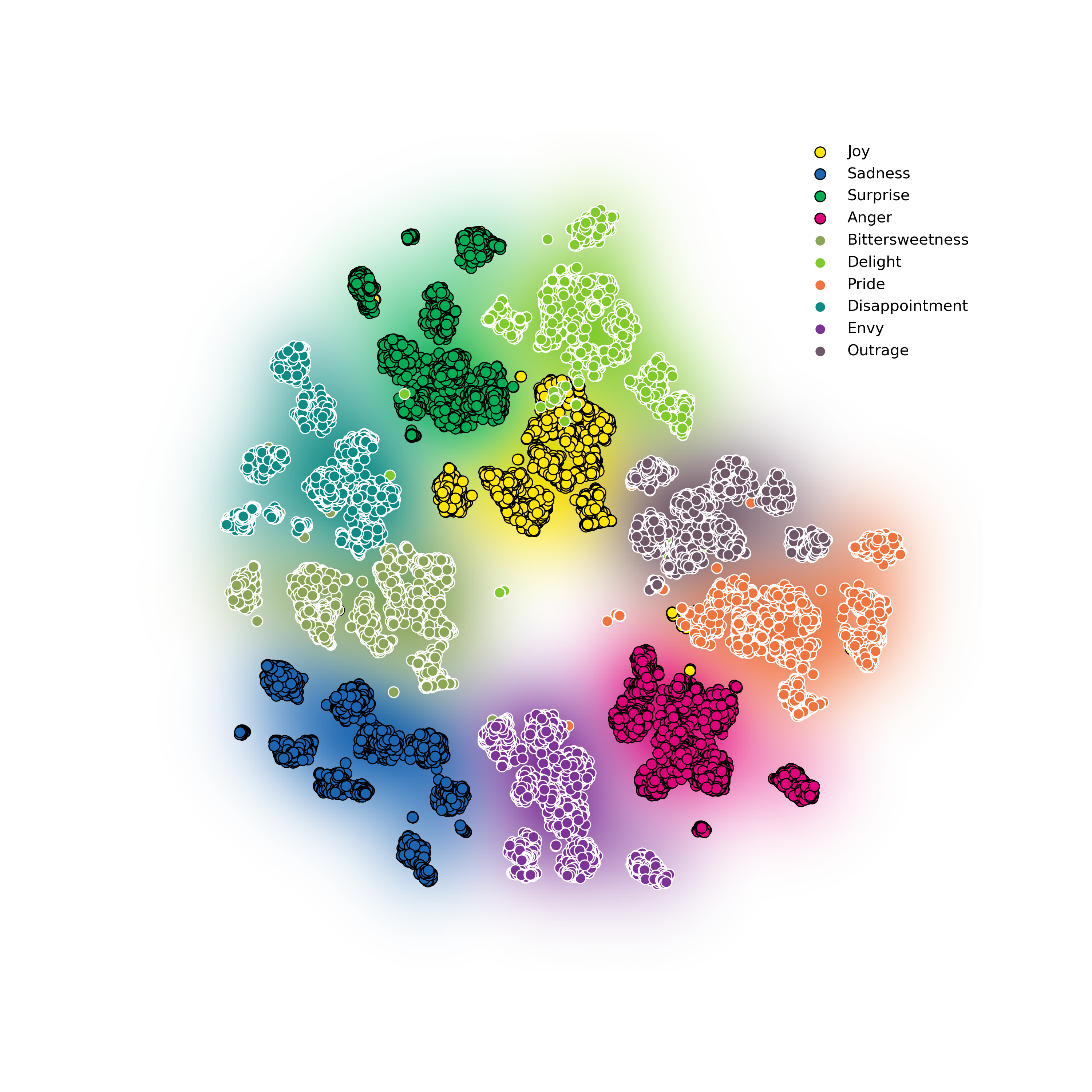}
    \caption{Emotionally-separable prosody embeddings learned from our proposed model, \textit{Daisy-TTS}. Emotions bordered in black denote primary emotions, while ones bordered in white denote secondary emotions derived from the mixture of primary ones.}
    \label{fig:primary-mixed-embs}
\end{figure} 

This intricate spectrum of emotions posits an interesting challenge in designing emotional text-to-speech systems. Most works have adapted psychological theories in their design to properly represent and simulate emotional speech, notably with the discrete~\citep{ekman1992argument} and dimensional~\citep{russell1980circumplex} theories of emotion. The former allows for a direct representation of emotions as discrete labels (e.g. \textit{anger}, \textit{joy}), but likely incapable of simulating other variety of emotions. The latter does open continuous emotional intensity simulation, but it reduces the emotion representations down into higher-level affect variables of \textit{valence}, \textit{arousal}, and \textit{dominance}. 

Another theory, the structural model of emotion~\citep{plutchik1984emotions}, represents a more structural view of emotions variety as derivative products of primary emotions. This would allow a wider representation of emotions, yet still underexplored in the emotional TTS design. To the best of our knowledge, it has only been adapted in a few neural emotional TTS, most notably~\citep{zhou2022speech,tang2023emomix}, where they managed to simulate intensity-level and mixture of emotions through rank-based method and emotion embedding conditioning, respectively.

Expanding upon prior works, we propose an emotional text-to-speech design–grounded on the structural model of emotion–from the perspective of prosody modelling. Our proposed design learns latent prosody embedding as a proxy for emotion, which not only improves the emotion simulation quality and perceivability of prior works, but also extends the ability to simulate a wider spectrum of emotion as theorized in the structural model. Our main contributions are summarized as follows:

\begin{itemize}[leftmargin=*]
    \item[$\circ$] We introduce an emotional text-to-speech design, Daisy-TTS, to simulate wide spectrum of emotion through learning \textbf{emotionally-separable prosody embeddings} (Figure~\ref{fig:primary-mixed-embs}).
    \item[$\circ$]  Through decomposing and manipulating the learned embeddings, we are capable of simulating: 
    \begin{itemize}[leftmargin=*]
        \item[\small{$\bullet$}] \textbf{Primary Emotions}, as learned from the training samples, e.g. \textit{joy}, \textit{sadness}, \textit{anger}.
        \item[\small{$\bullet$}] \textbf{Secondary Emotions}, as a mixture of primary emotions, e.g. \textit{envy} = \textit{sadness} + \textit{anger}.
        \item[\small{$\bullet$}] \textbf{Intensity-level}, by scaling the emotion embedding, e.g. \textit{rage} = 2 * \textit{anger}. 
        \item [\small{$\bullet$}] \textbf{Emotions Polarity}, by negating the emotion embedding, e.g. \textit{sadness} = \textit{- joy}.
    \end{itemize}
    \item[$\circ$] Our evaluation shows that our proposed design improves the quality of prior work~\citep{zhou2022speech} in terms of speech naturalness, emotion perceivability, and simulation capabilities.
\end{itemize}

\section{Related Work}

\subsection{Structural Model of Emotion} \label{sec:rw-smoe}
\begin{figure}[!htb]
    \centering
    \includegraphics[width=0.9\linewidth]{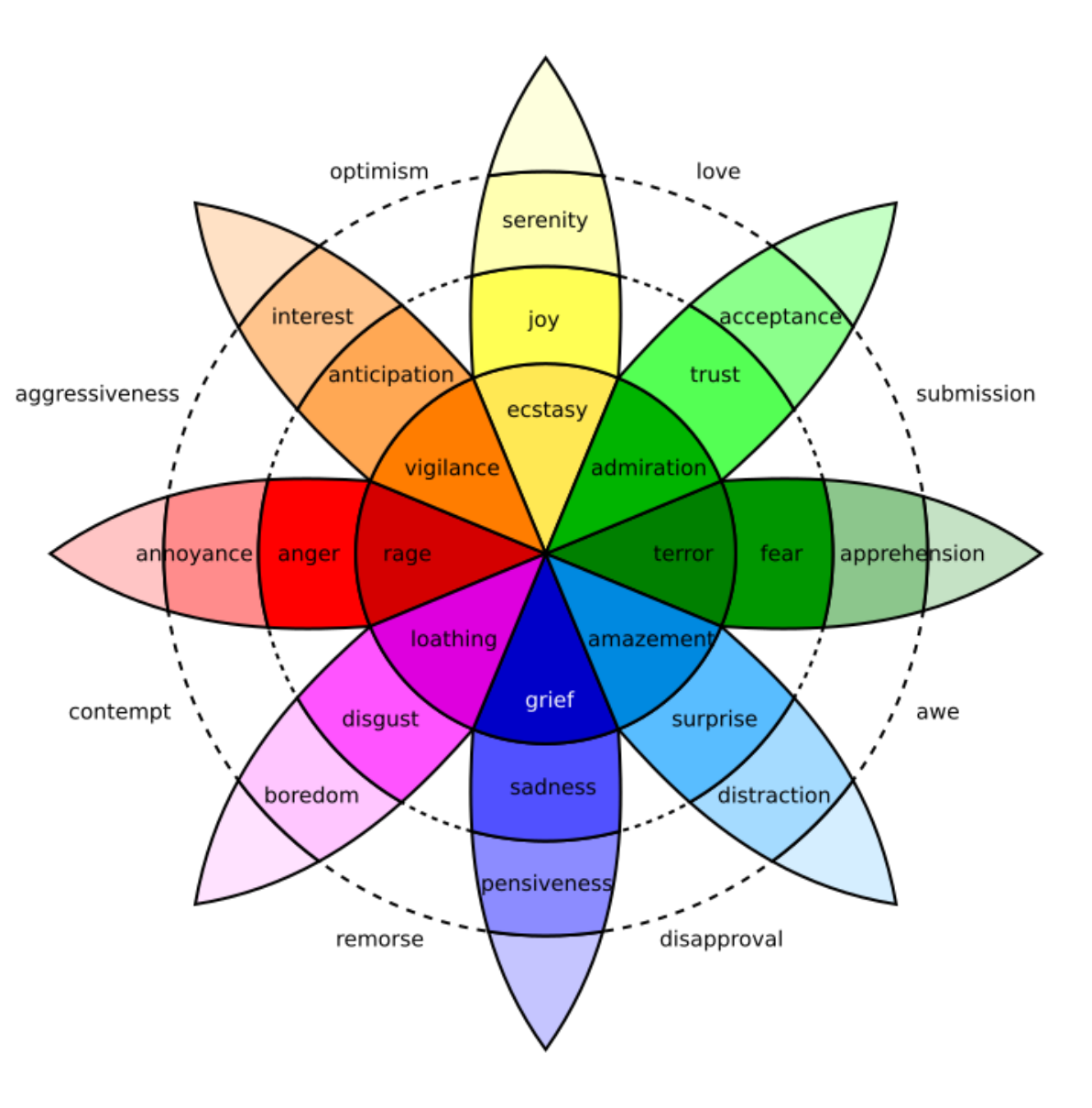}
    \caption{Visual Representation of the Structural Model of Emotions.}
    \label{fig:structural-model-visual}
\end{figure}

In the structural model of emotion~\citep{plutchik1982psychoevolutionary}, the sheer variety of emotional expressions is characterized as a derivative product of primary emotions. Figure~\ref{fig:structural-model-visual} shows a concise visualization of the model in a flower-like arrangement. 

Each petal represents the primary emotions. The length of the petal represent the emotion's intensity. The proximity of each petal represents the similarity and polarity of emotions to one another. And emotions in-between the petals represent the secondary emotions, derived from the mixture of primary emotions.
We found it quite appealing the way emotions are characterized in the structural model. It defined a discrete set of primary emotions, yet the characteristics of intensity, polarity, and mixing of emotions expand their representation beyond discrete labels.

Given this, we decided to adapt the structural model as the basis for modeling emotion. Specifically, we design our proposed emotion modelling framework with a focus on simulating emotions characterized by the structural model.


\subsection{Emotion Manifestation in Speech} \label{sec:manifest-emo}

It is intuitive to assume that speech conveys identifiable emotions, but what part of it plays a role? Speech is roughly composed of lexical (what we say) and non-lexical (how we say) components. Though both components do convey emotions~\citep{schuller2020reviewemo}, substantial works have shown that \textit{prosody}, a non-lexical pattern of tune, rhythm, and timbre, plays a main role in conveying emotion in speech~\citep{mozziconacci02_speechprosody,cowen2019primacy}. 

Based on this, we decided to incorporate prosody in our proposed emotion modelling framework, as a proxy representation of emotion in speech.

\subsection{Emotion Modelling in Text-to-Speech} \label{sec:rw-tts}

Modelling emotion in a text-to-speech task usually involves designing a conditional model that simulates emotional speech given text and emotion representation. How emotions are represented will affect the characteristics of emotions that can be simulated by the model. In~\citep{lee2017emotional}, emotion is represented as discrete labels (e.g. \textit{joy}, \textit{sadness}), though this allows the model to simulate primary emotions by explicitly specifying the input emotion label, it hinders the simulation of more complex characteristics such as emotion intensity and a mixture of emotions. To alleviate this, a few recent works adapt the mentioned structural model of emotions in their design.~\citep{zhou2022speech} is one of the firsts that introduce the ability to simulate emotion intensity and secondary emotion through a rank-based emotion attribute vector.~\citep{tang2023emomix} represents emotion as a vector embedding extracted from a pretrained speech emotion recognizer, which also allows the simulation of both characteristics by combining the hidden state of embedding.

The aforementioned models provide a solid framework for simulating a variety of emotion characteristics, but most of them haven't explicitly incorporated prosody modelling in their design, which as stated in Section~\ref{sec:manifest-emo}, prosody should have played a main role in conveying emotion in speech. One notable method to model prosody follows~\citep{skerryryan2018endtoend,wang2018style}, from which style embeddings are directly learned from speech. ~\citep{Za_di_2022} adopt this method to specifically learn cross-speaker prosody embedding. Though these methods were not explicitly designed to simulate characteristics of emotions,~\citep{Rijn_2021} shows that the embedding latent space of this family of models is reliably associated with primary emotions.

Based on this, we hypothesize that explicitly incorporating prosody modelling to model emotions--or framing the latter as the former--would allow us to simulate a wider range of emotional characteristics.

\begin{figure*}
    \centering
    \includegraphics[trim={2cm 1cm 2cm 2cm},width=0.8\linewidth]{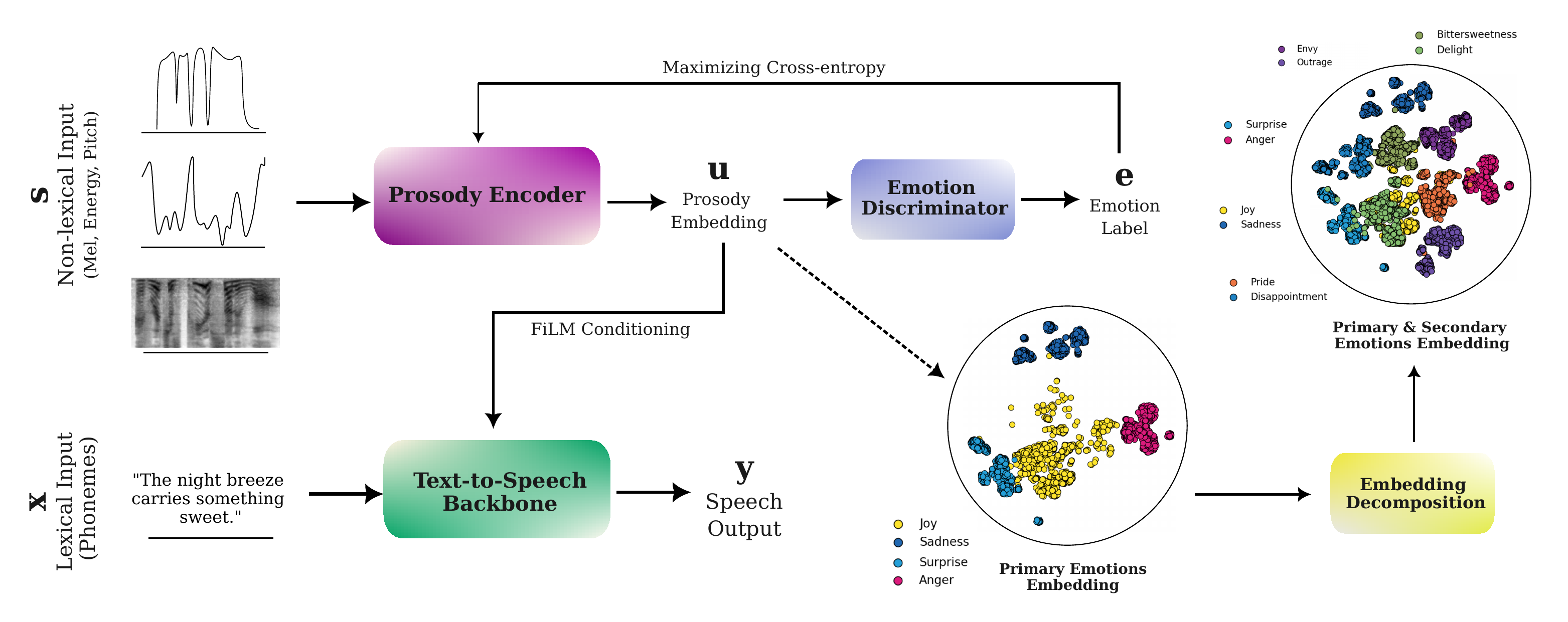}
    \caption{Systemic Overview of Daisy-TTS. Emotionally-separable prosody embeddings were learned from a set of speech features and used to condition a TTS backbone model. To simulate a wider range of emotion characteristics, such as intensity, polarity, and mixture of emotions, an embedding decomposition was applied to the learned embeddings.}
    \label{fig:daisy-illustration}
\end{figure*}

\section{Learning Emotionally-separable Prosody Embedding}

In this section, we discuss our approach to learning emotionally-separable prosody embeddings through our proposed framework, \textit{Daisy-TTS}. As in Figure~\ref{fig:daisy-illustration}, the embeddings were learned by encoding non-lexical speech features through a prosody encoder equipped with an emotion discriminator and conditioned them to a text-to-speech backbone.

\subsection{Text-to-Speech Backbone}

Let $\mathcal{F}(x|u)$ be the backbone TTS model to synthesize prosody-controlled speech, $y$, given lexical and prosodic features, $\{x,u\}$. We represent $x$ and $y$ as phonemes and mel-spectrogram. While for $u$, we assume a vector embedding representing prosody, $u \in \mathbb{R}^{d}$, with $d$ as the size of the embedding.

We chose Grad-TTS~\citep{popov2021gradtts} as our backbone. Grad-TTS is a non-autoregressive TTS model based on diffusion probabilistic modelling. We chose this model as it is one the most high-quality open-source TTS at the time of this research, which allow us to focus more on the emotion modelling aspect.

In our implementation, we follow the original architecture of Grad-TTS, which is composed of encoder and decoder modules. We apply a slight modification to the encoder for prosody conditioning. The encoder follows the same structure as~\citep{kim2020glowtts}, consisting of a 1D convolutional prenet, 6 Transformer blocks, and a linear projection layer. The encoder takes in $x$ as direct input, $u$ is conditioned into the module on feature-level with FiLM conditioning~\citep{perez2017film}, the module outputs $\mu_{\{x,u\}}$, hidden feature of $\{x,u\}$ assumed to be aligned with corresponding speech, $y$. The decoder follows the same U-Net architecture as in~\citep{ronneberger2015unet,ho2020denoising}. It synthesizes $y$ by iteratively decoding a Gaussian noise, $z \sim \mathcal{N}(\mu_{\{x,u\}},I)$, parametrized by the encoder's output. 

\subsection{Emotionally-separable Prosody Encoder} \label{sec:prosody-encoder}

To obtain a vector embedding representation of prosody, $u$, we follow the methods from prosody transfer modelling~\citep{skerryryan2018endtoend,wang2018style,Za_di_2022}. We define $\mathcal{G}(s)$, a prosody encoder model independent of the backbone model. The main objective of $\mathcal{G}(s)$ is to learn an emotionally-separable prosody embedding, $u$, from some non-lexical features, $s$. 

We choose multiple non-lexical features, which are mel-spectrogram, pitch contour, and energy contour, to provide rich bases to encode prosodic information. $\mathcal{G}(s)$ takes in $s$, and gradually collapse the temporal dimension of the features. By collapsing the temporal dimension, we assume it will also collapse the heavily temporal-dependent information in the feature as well, such as the lexical features, which are still present in the mel-spectrogram. 

To impose emotion separability in the prosody embedding space, we introduce an auxiliary emotion discriminator at the end of $\mathcal{G}(s)$. Our empirical findings, as illustrated in Section~\ref{sec:ablation-disc}, shows that by maximizing the cross entropy between prosody embeddings and their respective emotion labels can effectively impose emotion separability of the prosody embeddings.


\subsection{Model Training} \label{sec:daisy-training}

\textbf{Training Objective.} Both the backbone and the prosody encoder are trained jointly. There are 2 main training objectives, each corresponding to both models' respective tasks in synthesizing prosody-controlled speech and imposing emotion separability. For the former, we follow the same training objectives in~\citep{popov2021gradtts}, and for the latter, we maximize the cross-entropy loss between the prosody embeddings and their respective emotion labels.

\vspace{2mm}

\noindent\textbf{Training Dataset.} We found ESD (Emotional Speech Dataset) \citep{zhou2021seen,zhou2022emotional} is suitable for our problem. We select the subset containing 10 English speakers (5 male, 5 female), each with 350 parallel utterances voice acted in neutral and 4 primary emotions: \textit{joy}, \textit{sadness}, \textit{anger}, and \textit{surprise}. We follow the original training samples split. With data processing protocol follows \citep{popov2021gradtts}, where the texts are converted to phonemes and the speech samples are represented as mel-spectrograms with a sample rate set to 22,050 Hz, and mel bands, filter size, window size, and hop length set to 80, 1024, 1024, 256, respectively.

\vspace{2mm}

\noindent\textbf{Training Setting.} Our model parameters count is \~14,800,000. We trained the model in a single GPU (NVIDIA A100). AdamW~\citep{loshchilov2019decoupled} is used as the optimizer, with $\beta_{1} = 0.9$, $\beta_{2} = 0.99$, and constant learning rate of $1e^{-4}$. The model was trained for 310,000 steps with a batch size of 32.

\noindent\textbf{Vocoder Setting.} As most neural TTS that use mel-spectrogram as the output, we need a vocoder to decode the synthesized speech. We use the pretrained HifiGAN~\citep{kong2020hifigan} vocoder available in the official repository of Grad-TTS~\footnote{\url{https://github.com/huawei-noah/Speech-Backbones/tree/main/Grad-TTS}}, and finetuned it with ESD dataset.


\section{Simulating Emotion through Prosody Embedding} \label{sec:emo-simulate}

After Daisy-TTS is trained, we infer a set of prosody embeddings from training speech samples. We denote $\textbf{u} = \{\mathcal{G}(s_{i})\}_{i=1}^{m} \in \mathbb{R}^{m \times d}$, as the inferred embeddings, where $m$ is the total number of samples. As each $s_{i}$ is labeled with primary emotions $\varepsilon$ = \{\textit{joy}, \textit{sadness}, \textit{anger}, \textit{surprise}\}, we have $\textbf{u}^{\varepsilon} \subset \textbf{u}$.

\subsection{Emotion Separability}

We examine the emotion separability of $\textbf{u}$ by applying the dimensionality reduction method, t-SNE~\citep{van2008visualizing}. Figure~\ref{fig:primary-mixed-embs} shows how each $\textbf{u}^{\varepsilon}$ is well separated from each other. This implies that the encoded information describing primary emotion in $\textbf{u}$ are distinguishable from one another, which conforms to the first emotion characteristics described in Section~\ref{sec:rw-smoe}. 

This separability is essential to simulate secondary and polar emotions, as both are assumed to be derived from independent primary emotions.


\subsection{Emotion from Prosody Decomposition}

As we want to simulate a range of emotions through $u$, it is important to have controllability over $u$ beyond providing an exemplar speech to the model. Our approach is inspired by~\citep{blanz2023morphable}. We decompose $u$ as a linear combination of its "prototypes" weighted by parameter vector $\textbf{w}$:
\begin{equation} \label{eq:prosody-decomp}
u(\textbf{w}) = \overline{u} + \sum_{i=1}^{n} w_{i}v_{i}
\end{equation}

the above decomposition can be obtained by performing PCA~\citep{wold1987principal} to the embeddings, with $\overline{u}$ as the mean of embeddings, $n$ as the chosen number of principal components, $v_{i} \in \mathbb{R}^{n \times d}$ as the embedding prototypes in the form of its top-$n$ eigenvectors, and $\textbf{w}$ as the parameter vector associated with $u$.

This type of decomposition assumed that $\textbf{w}$ follow a multivariate Gaussian distribution~\citep{egger20203d}, with probability density function described as:
\begin{equation} \label{eq:w-dist}
p(\textbf{w}) \sim e^{-\frac{1}{2}\sum_{i=1}^{n} {(w_{i}/\sigma_{i})^{2}}}
, \, \textbf{w} \sim \mathcal{N}(0, \Sigma)
\end{equation}
where $\sigma_{i}^{2}$ being the eigenvalues correspond to $v_{i}$, and $\Sigma$ is the covariance matrix with diagonals of $\sigma_{i}$. With this, we can synthesize plausible prosody embedding conveying certain emotions by sampling $\textbf{w}$ from the distribution.

\subsection{Simulating Primary Emotions}

We can simulate primary emotions by constraining the synthesis of $u(\textbf{w})$ into a certain class of $\varepsilon$. To do this, we sample $\textbf{w}$ that is centered around $\textbf{u}^{\varepsilon}$, denoted as $\textbf{w}^{\varepsilon}$. The mean of the distribution, $\mu^{\varepsilon}_{i}$, can be interpreted as the principal components of $\overline{\textbf{u}}_{\varepsilon}$. The density function of $\textbf{w}^{\varepsilon}$ is defined as:
\begin{equation}
p(\textbf{w}^{\varepsilon}) \sim e^{-\frac{1}{2}\sum_{i=1}^{n} { (w^{\varepsilon}_{i}-\mu^{\varepsilon}_{i}})^{2} / \sigma_{i}^{2}  }
, \, \textbf{w}^{\varepsilon} \sim \mathcal{N}(\mu^{\varepsilon}_{i}, \Sigma)
\end{equation}

We can then sample a plausible $\textbf{w}^{\varepsilon}$ to synthesize prosody embedding that simulates the desired primary emotion, $\varepsilon$.

\subsection{Simulating Secondary Emotions}

A secondary emotion is defined as a mixture between 2 primary emotions~\citep{plutchik1982psychoevolutionary}. With our prosody decomposition, we can interpret a secondary emotion as a mixture between two different $\textbf{w}^{\varepsilon}$, denoted as $\textbf{w}^{\varepsilon_{s}}$. As $\textbf{w}$ follow multivariate Gaussian distribution, we can define the mean and covariance for $\textbf{w}^{\varepsilon_{s}}$ as:
\begin{equation}
\Sigma^{\varepsilon_{s}} = (2\Sigma^{-1})^{-1}
\end{equation}
\begin{equation}
\mu^{\varepsilon_{s}} = \Sigma^{\varepsilon_{s}}\Sigma^{-1}\mu^{{\varepsilon_{1}}} + \Sigma^{\varepsilon_{s}}\Sigma^{-1}\mu^{{\varepsilon_{2}}}
\end{equation}
From this, we can sample $\textbf{w}^{\varepsilon_{s}}$ to synthesize prosody embedding conveying a variety of secondary emotions:
\begin{equation}
\textbf{w}^{\varepsilon_{s}} \sim \mathcal{N}(\mu^{\varepsilon_{s}}, \Sigma^{\varepsilon_{s}})
\end{equation}

Figure~\ref{fig:primary-mixed-embs} shows the projected embeddings of secondary emotions. We could see that each secondary emotion lies between the approximate area of its primary emotion pairs. For instance, \textit{bittersweetness} lies between the area of \textit{joy} and \textit{sadness}, \textit{envy} lies between the area of \textit{sadness} and \textit{anger}, and so on.

\subsection{Simulating Intensity-level}
We found that performing a scaling operation on the basis with scalar, $\alpha \sum_{i=1}^{n} w_{i}v_{i}$, equates to simulating the intensity-level of emotion expressed by $u(\textbf{w})$. With $\alpha < 1.0$ equates to lowering the intensity of the emotion, while $\alpha > 1.0$ intensifies it. 

\subsection{Emotions Polarity}

Polarity in emotion is defined as the maximum dissimilarity of emotion~\citep{plutchik1982psychoevolutionary}. We found that by negating $\textbf{w}^{\varepsilon}$, we can simulate an emotion that is perceivably the opposite of that emotion. For example, sampling $-\textbf{w}^{\varepsilon_{joy}}$ gives embedding that is similar to sampling from $\textbf{w}^{\varepsilon_{sadness}}$ and vice versa. 





\section{Evaluation}

In this section, we report the evaluation we conducted on the model's capability to simulate emotions and how humans subjectively perceived them.

\subsection{Evaluation Metrics} \label{sec:eval-metrics}

\begin{table*}[!htb]
    \small\centering
    \begin{tabular}{@{}l@{ ~ }cccccc@{ ~ }c}
\toprule
 & \multicolumn{7}{c}{Speech Naturalness (MOS; 95\% CI) $\uparrow$}\\
 \midrule
         &  \color{teal}\textit{Anger}&  \color{teal}\textit{Joy}&  \color{teal}\textit{Sadness}&  \color{teal}\textit{Surprise}& \color{violet}\textit{Delight} & \color{violet}\textit{Pride}& \color{violet}\textit{Envy} \\
\midrule
         Ground Truth&  3.844$_{\pm 0.129}$&  3.867$_{\pm 0.123}$&   3.567$_{\pm 0.133}$&  3.789$_{\pm 0.133}$&  -& -& -\\
         Baseline&  3.356$_{\pm 0.189}$&  3.167$_{\pm 0.201}$&  3.200$_{\pm 0.186}$&   2.822$_{\pm 0.246}$&   3.178$_{\pm 0.209}$& 3.356$_{\pm 0.204}$& 3.478$_{\pm 0.186}$\\
 Daisy-TTS& \textbf{3.689}$_{\pm 0.192}$& \textbf{3.844}$_{\pm 0.180}$&  \textbf{3.456}$_{\pm 0.196}$& \textbf{3.511}$_{\pm 0.186}$& \textbf{3.811}$_{\pm 0.182}$& \textbf{3.656}$_{\pm 0.166}$&  \textbf{3.900}$_{\pm 0.172}$\\
 \midrule
 & \multicolumn{7}{c}{Emotion Perceivability (Acc.) $\uparrow$}\\
 \midrule
 & \color{teal}\textit{Anger} & \color{teal}\textit{Joy} & \color{teal}\textit{Sadness} & \color{teal}\textit{Surprise}& \color{violet}\textit{Delight} & \color{violet}\textit{Pride} & \color{violet}\textit{Envy} \\
 \midrule
 Ground Truth& 0.656& 0.478& 0.556& 0.489& -& -& -\\
 Baseline& 0.300& 0.233& \textbf{0.533}& 0.144& 0.311& 0.133& 0.255\\
 Daisy-TTS& \textbf{0.533}& \textbf{0.333}& 0.478& \textbf{0.300}& \textbf{0.355}& \textbf{0.166}& \textbf{0.277}\\
 \toprule
    \end{tabular}
    \caption{Results of MOS and emotion perception tests on {\color{teal}primary} and {\color{violet}secondary} emotions (continued in Appendix~\ref{sec:app-eval-result}). Daisy-TTS overall achieves higher speech naturalness and emotion perceivability compared to the baseline.}
    \label{tab:primary-secondary-emo-mos}
\end{table*}

\begin{figure*}[!htb]
    \centering
    \includegraphics[trim={5cm 0.5cm 5cm 1.5cm}, width=0.95\linewidth]{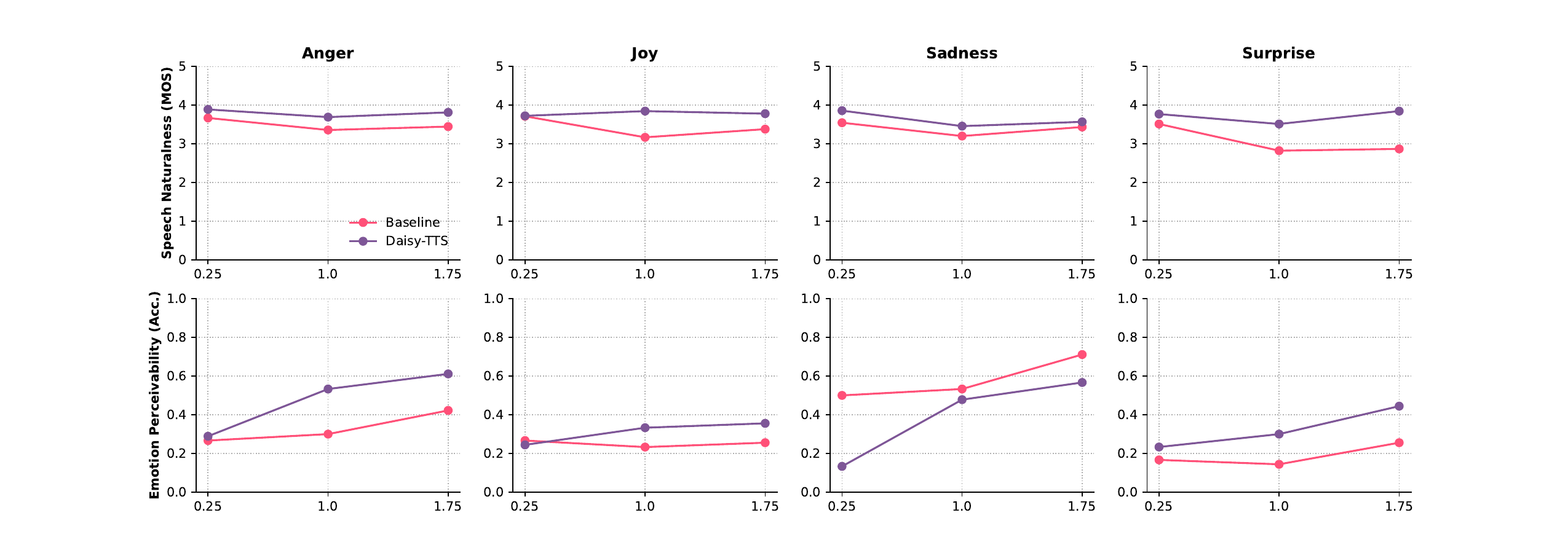}
    \caption{Result of emotion perception test for different intensity-level of primary emotions.}
    \label{fig:primary-intensity-plot}
\end{figure*}

We adopt the tests for emotional speech evaluation conducted in~\citep{zhou2022speech,cowen2019primacy}. Emotional speech samples will be evaluated in terms of its speech naturalness and emotion perceivability. Speech naturalness is evaluated through the MOS (Mean Opinion Score) test, where human annotators were asked to rate the naturalness of the given speech samples on a 5-point Likert scale: Excellent, Good, Fair, Poor, and Bad. Whereas emotion perceivability is evaluated through an emotion perception test, where the annotators were asked to select one or two emotions they perceived.

To accommodate the test, we crowd-sourced our evaluation to human annotators. The details of the crowd-sourcing setup can be found in~\ref{sec:mturk-setup}

\subsection{Baseline Setup}

As discussed in Section~\ref{sec:rw-tts}, there are mainly 2 candidates for our baselines, ~\citep{zhou2022speech} and~\citep{tang2023emomix}. We can only utilize~\citep{zhou2022speech} as our baseline as the latter is not open-sourced.

Through their rank-based method, the baseline model can simulate primary, secondary, and intensity-level of emotions. The model is a sequence-to-sequence TTS akin to the recognition-synthesis model \citep{Zhang_2020}. We use the official pretrained model shared by the authors~\footnote{\url{https://github.com/KunZhou9646/Mixed_Emotions}}. 

It is important to note that the shared model was only trained on a single English speaker subset of ESD (\textit{speaker "0019"}), the audio configuration is not the same as described in Section~\ref{sec:daisy-training}, and there was no pretrained vocoder to decode the mel-spectrogram. Thus, to make the evaluation fairer, we only compare the samples spoken by \textit{speaker "0019"} and we also finetune the same vocoder we described in Section~\ref{sec:daisy-training} with ESD dataset but follow the audio configuration of the baseline. 

\subsection{Evaluation Results}

\subsubsection{Primary Emotions Evaluation} \label{sec:primary-eval}
Results are shown in Table~\ref{tab:primary-secondary-emo-mos}. In MOS test, Daisy-TTS outperformed the baseline across all 4 primary emotions. In the emotion perception test, Daisy-TTS achieves higher overall emotion perceiveability compared to the baseline, with an exception only in expressing \textit{sadness}.

Interestingly, if we observe the ground truth's emotion perceivability, it shows that the ground-truth emotions are not easily distinguishable by humans either. This actually conforms with recent insights in~\citep{troiano2021emotion} on the frequent inconfidence of humans in rating emotional tasks. We will further discuss this in Section~\ref{sec:confusion}. 

\subsubsection{Secondary Emotion Evaluation}
As we don't have ground truth samples for secondary emotions, we conducted the tests only with Daisy-TTS and the baseline. Results are shown in Table~\ref{tab:primary-secondary-emo-mos}. Similar to the previous results, Daisy-TTS overall outperforms the baseline, with the exception of expressing \textit{disappointment}, a mixture of \textit{surprise} and \textit{sadness}, which can be explained by the baseline's stronger ability to express \textit{sadness}.

\subsubsection{Intensity-level Evaluation} \label{sec:intensity-eval}
We simulate 3 different intensity levels: low, medium, and high intensity, respectively represented by $\alpha = \{{0.25, 1.0, 1.75\}}$. Figure~\ref{fig:primary-intensity-plot} shows how different intensity levels affect the naturalness and perceivability of primary emotions. MOS varies when adjusting the intensity and we see occasional drop. However, our method is consistently outperforms the baseline. The baseline suffered from the highest drop of 0.689, while Daisy-TTS highest drop is only 0.255. This implies that Daisy-TTS is more robust in expressing intensity-varying emotions compared to the baseline. 

In emotion perception test, we observe that by increasing the intensity-level, the emotion perceivability is improved as well, with Daisy-TTS achieves higher improvement across emotions except for \textit{sadness}, which conforms to previous tests. This behavior can be explained intuitively from the assumption of emotions would be perceived more confidently if expressed with stronger intensity~\citep{troiano2021emotion}.

\subsubsection{Emotions Polarity Evaluation}
As Daisy-TTS is the only model in the evaluation capable of simulating emotions polarity, we only conducted the evaluation with the model. Results are shown in Table~\ref{tab:mos-emo-polar}. Simulated polar version of \textit{joy} seems to be as perceivable as its \textit{sadness} counterpart, while the perceiveability of the polar version of \textit{sadness} has a 17.8\% drop from \textit{joy}.  
\begin{table}[]
\small
\centering
\begin{tabular}{cc|cc} 
\toprule 
 \multicolumn{4}{c}{Speech Naturalness (MOS) $\uparrow$}\\
\midrule
  \textit{Joy}& \textit{Polar Sadness}& \textit{Sadness}&\textit{Polar Joy}\\
\midrule
 3.844& 3.666$_{\textcolor{red}{-.177}}$&3.456&3.788$_{\textcolor{teal}{+.331}}$\\
\midrule
  \multicolumn{4}{c}{Emotion Perceivability (Acc.) $\uparrow$}\\
  \midrule
 \textit{Joy}& \textit{Polar Sadness}& \textit{Sadness}&\textit{Polar Joy}\\
 \midrule
  0.333& 0.344$_{\textcolor{teal}{+.011}}$& 0.478&0.300$_{\textcolor{red}{-.178}}$\\
\midrule
\end{tabular}
\caption{Results of MOS and emotion perception tests of Daisy-TTS from simulating emotions polarity.}
\label{tab:mos-emo-polar}
\end{table}

\section{Discussion} \label{sec:discussion}

\subsection{Human Confusion across Primary Emotions} \label{sec:confusion}

As reported in Section~\ref{sec:primary-eval}, we found it intriguing that even the emotions in ground truth samples are not easily distinguishable by humans either. But, we could actually explain this from two perspectives. First, as described in Section~\ref{sec:rw-smoe}, emotions may vary in their degree of similarity to one another, thus some emotions may be perceived similarly and confused with the others. Second, based on the findings in Section~\ref{sec:intensity-eval} and~\citep{troiano2021emotion}, it might also be due to the intensity-level of the evaluated samples. 

We can further examine this by looking at the cosine similarity for each primary emotion embedding. Table~\ref{fig:cosine-matrix-primary} shows that most emotions are not fully dissimilar to one another, there is still some degree of similarity across emotions, even for polar opposites like \textit{joy} and \textit{sadness}.
\begin{table}[!htb]
\small
\centering
\begin{tabular}{lllll}
           & \multicolumn{1}{l}{Sadness}                          & \multicolumn{1}{l}{Joy}                              & \multicolumn{1}{l}{Anger}                           & \multicolumn{1}{l}{Surprise}                         \\
  Sadness  & \cellcolor[HTML]{432371}{\color[HTML]{FFFFFF} +1.0}  & \cellcolor[HTML]{D5CEDF}-0.46                        & \cellcolor[HTML]{D5CEDF}-0.43                       & \cellcolor[HTML]{D5CEDF}{\color[HTML]{333333} -0.40} \\
  Joy      & \cellcolor[HTML]{D5CEDF}-0.46                        & \cellcolor[HTML]{432371}{\color[HTML]{FFFFFF} +1.0}  & \cellcolor[HTML]{AB9DC0}-0.15                       & \cellcolor[HTML]{AB9DC0}{\color[HTML]{333333} +0.20} \\
  Anger    & \cellcolor[HTML]{D5CEDF}-0.43                        & \cellcolor[HTML]{AB9DC0}-0.15                        & \cellcolor[HTML]{432371}{\color[HTML]{FFFFFF} +1.0} & \cellcolor[HTML]{EAE7EF}-0.56                        \\
  Surprise & \cellcolor[HTML]{D5CEDF}{\color[HTML]{333333} -0.40} & \cellcolor[HTML]{AB9DC0}{\color[HTML]{333333} +0.20} & \cellcolor[HTML]{EAE7EF}-0.56                       & \cellcolor[HTML]{432371}{\color[HTML]{FFFFFF} +1.0} 
\end{tabular}
\caption{Cosine Similarity between Primary Emotion Embeddings.}
\label{fig:cosine-matrix-primary}

\begin{tabular}{llllll}
 &          & \multicolumn{4}{c}{Predicted Label}                                                                                                                                                                                   \\
 &          & \multicolumn{1}{c}{Sadness}                         & \multicolumn{1}{c}{Joy}                             & \multicolumn{1}{c}{Anger}                           & \multicolumn{1}{c}{Surprise}                        \\
 \parbox[t]{2mm}{\multirow{4}{*}{\rotatebox[origin=c]{90}{True Label}}}& Sadness  & \cellcolor[HTML]{6D5491}{\color[HTML]{FFFFFF} 0.56} & \cellcolor[HTML]{EAE7EF}{\color[HTML]{000000} 0.21} & \cellcolor[HTML]{FFFFFF}{\color[HTML]{000000} 0.04} & \cellcolor[HTML]{EAE7EF}{\color[HTML]{000000} 0.19} \\
 & Joy      & \cellcolor[HTML]{FFFFFF}{\color[HTML]{000000} 0.04} & \cellcolor[HTML]{AB9DC0}{\color[HTML]{000000} 0.48} & \cellcolor[HTML]{D5CEDF}{\color[HTML]{000000} 0.28} & \cellcolor[HTML]{EAE7EF}{\color[HTML]{000000} 0.2}  \\
 & Anger    & \cellcolor[HTML]{FFFFFF}{\color[HTML]{000000} 0.05} & \cellcolor[HTML]{EAE7EF}{\color[HTML]{000000} 0.2}  & \cellcolor[HTML]{6D5491}{\color[HTML]{FFFFFF} 0.66} & \cellcolor[HTML]{FFFFFF}{\color[HTML]{000000} 0.08} \\
 & Surprise & \cellcolor[HTML]{FFFFFF}{\color[HTML]{000000} 0.03} & \cellcolor[HTML]{D5CEDF}{\color[HTML]{000000} 0.32} & \cellcolor[HTML]{EAE7EF}{\color[HTML]{000000} 0.16} & \cellcolor[HTML]{AB9DC0}{\color[HTML]{000000} 0.49}
\end{tabular}
\caption{Confusion matrix of ground truth.}
\label{fig:confusion-matrix-truth}

\begin{tabular}{llllll}
 &          & \multicolumn{4}{c}{Predicted Label}                                                                                                                                                                                   \\
 &          & \multicolumn{1}{c}{Sadness}                         & \multicolumn{1}{c}{Joy}                             & \multicolumn{1}{c}{Anger}                           & \multicolumn{1}{c}{Surprise}                        \\
 \parbox[t]{2mm}{\multirow{4}{*}{\rotatebox[origin=c]{90}{True Label}}}& Sadness  & \cellcolor[HTML]{AB9DC0}{\color[HTML]{000000} 0.53} & \cellcolor[HTML]{D5CEDF}{\color[HTML]{000000} 0.27} & \cellcolor[HTML]{FFFFFF}{\color[HTML]{000000} 0.06} & \cellcolor[HTML]{EAE7EF}{\color[HTML]{000000} 0.13} \\
 & Joy      & \cellcolor[HTML]{C0B6D0}{\color[HTML]{000000} 0.42} & \cellcolor[HTML]{D5CEDF}{\color[HTML]{000000} 0.23} & \cellcolor[HTML]{EAE7EF}{\color[HTML]{000000} 0.21} & \cellcolor[HTML]{EAE7EF}{\color[HTML]{000000} 0.13} \\
 & Anger    & \cellcolor[HTML]{D5CEDF}{\color[HTML]{000000} 0.32} & \cellcolor[HTML]{C0B6D0}{\color[HTML]{000000} 0.34} & \cellcolor[HTML]{D5CEDF}{\color[HTML]{000000} 0.30} & \cellcolor[HTML]{FFFFFF}{\color[HTML]{000000} 0.03} \\
 & Surprise & \cellcolor[HTML]{C0B6D0}{\color[HTML]{000000} 0.33} & \cellcolor[HTML]{D5CEDF}{\color[HTML]{000000} 0.30} & \cellcolor[HTML]{D5CEDF}{\color[HTML]{000000} 0.22} & \cellcolor[HTML]{EAE7EF}{\color[HTML]{000000} 0.14}
\end{tabular}
\caption{Confusion Matrix of Baseline.}
\label{fig:confusion-matrix-baseline}

\begin{tabular}{llllll}
 &          & \multicolumn{4}{c}{Predicted Label}                                                                                                                                                                                   \\
 &          & \multicolumn{1}{c}{Sadness}                         & \multicolumn{1}{c}{Joy}                             & \multicolumn{1}{c}{Anger}                           & \multicolumn{1}{c}{Surprise}                        \\
 \parbox[t]{2mm}{\multirow{4}{*}{\rotatebox[origin=c]{90}{True Label}}}& Sadness  & \cellcolor[HTML]{AB9DC0}{\color[HTML]{000000} 0.48} & \cellcolor[HTML]{D5CEDF}{\color[HTML]{000000} 0.32} & \cellcolor[HTML]{FFFFFF}{\color[HTML]{000000} 0.08} & \cellcolor[HTML]{EAE7EF}{\color[HTML]{000000} 0.11} \\
 & Joy      & \cellcolor[HTML]{EAE7EF}{\color[HTML]{000000} 0.16} & \cellcolor[HTML]{C0B6D0}{\color[HTML]{000000} 0.33} & \cellcolor[HTML]{D5CEDF}{\color[HTML]{000000} 0.30} & \cellcolor[HTML]{EAE7EF}{\color[HTML]{000000} 0.21} \\
 & Anger    & \cellcolor[HTML]{EAE7EF}{\color[HTML]{000000} 0.11} & \cellcolor[HTML]{D5CEDF}{\color[HTML]{000000} 0.30} & \cellcolor[HTML]{AB9DC0}{\color[HTML]{000000} 0.53} & \cellcolor[HTML]{FFFFFF}{\color[HTML]{000000} 0.05} \\
 & Surprise & \cellcolor[HTML]{FFFFFF}{\color[HTML]{000000} 0.06} & \cellcolor[HTML]{C0B6D0}{\color[HTML]{000000} 0.36} & \cellcolor[HTML]{D5CEDF}{\color[HTML]{000000} 0.28} & \cellcolor[HTML]{D5CEDF}{\color[HTML]{000000} 0.30}
\end{tabular}
\caption{Confusion Matrix of Daisy-TTS.}
\label{fig:confusion-matrix-daisy}
\end{table}
This similarity is reflected by the confusion human annotators made as well. In Table~\ref{fig:confusion-matrix-truth}, we can see that the low perceivability for ground-truth samples might be attributed to the misperception of other emotions. For example, some expression of \textit{sadness} is mistaken for \textit{joy} even though they are polar opposites, \textit{joy} was often mistaken for \textit{anger} or \textit{surprise}, and vice versa. Table~\ref{fig:confusion-matrix-baseline} and~\ref{fig:confusion-matrix-daisy} shows the confusion in the models samples, the baseline seems to indicate higher misperception of primary emotions compared to the ground-truth. While Daisy-TTS seems to produce emotional speech that is perceived more similarly to the ground truth samples.

\subsection{Role of Emotion Discriminator} \label{sec:ablation-disc}
As our method to simulate emotions relies on the emotion separability of the prosody embeddings, we examine whether training with emotion discriminator actually affects the embeddings. 

\begin{figure}[!htb]
    \centering
    \includegraphics[trim={1cm 0cm 1cm 1cm},width=1\linewidth]{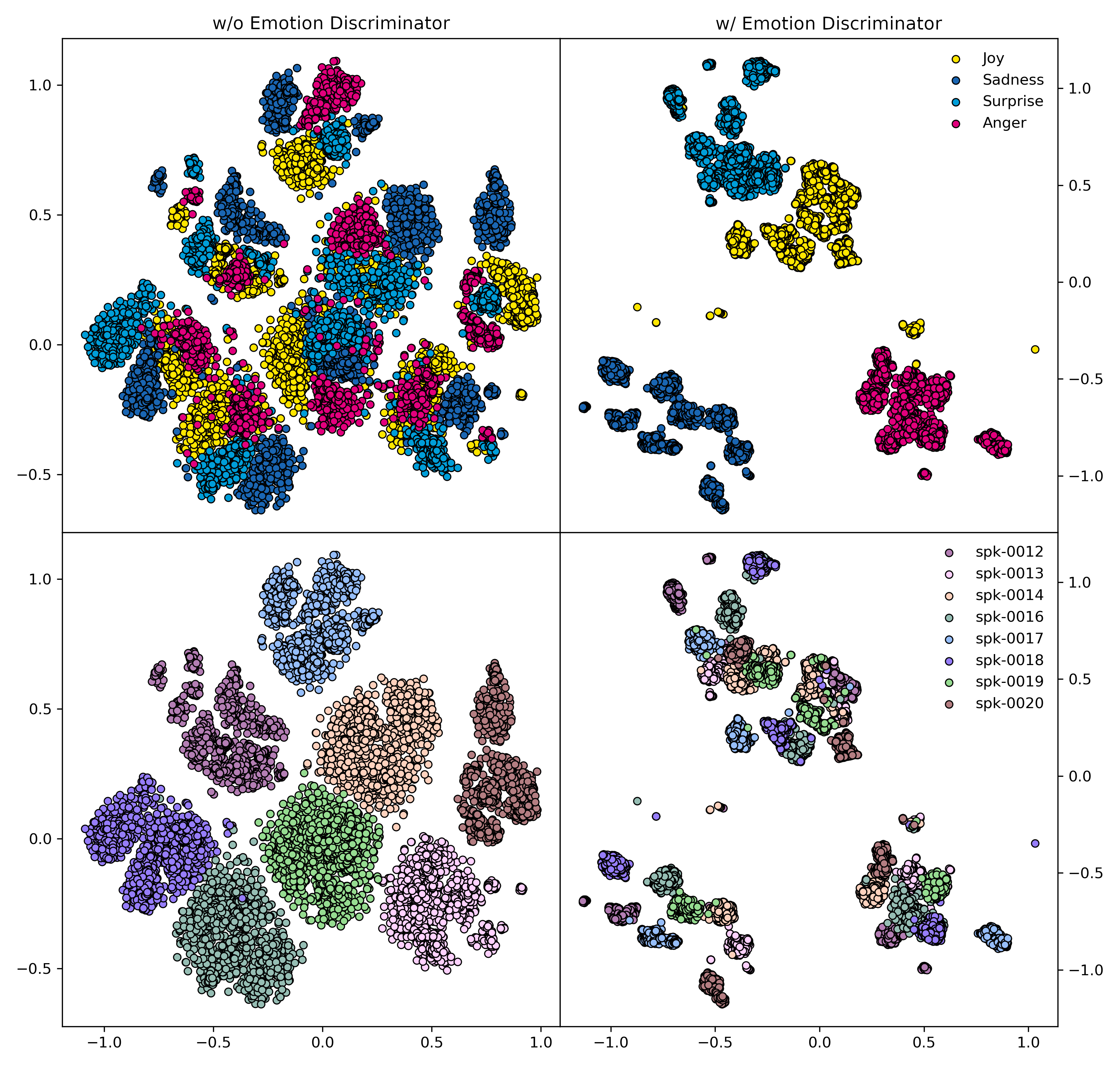}
    \caption{(Left Column) Prosody embeddings from training Daisy-TTS without emotion discriminator; (Right Column) with emotion discriminator. Training without emotion discriminator discouraged emotion separability.}
    \label{fig:ablation-embs-compare}
\end{figure}

Figure~\ref{fig:ablation-embs-compare} shows the comparison between the embeddings with and without the emotion discriminator. Absence of emotion discriminator does make the embedding space not emotionally-separable. However, it turns out that the embeddings would be separated by the speaker's identity. But, this speaker separability is retained in model with emotion discriminator, albeit locally in each emotion cluster.

\begin{figure}[!htb]
    \centering
    \includegraphics[trim={2cm 0cm 2.5cm 1.5cm},width=1\linewidth]{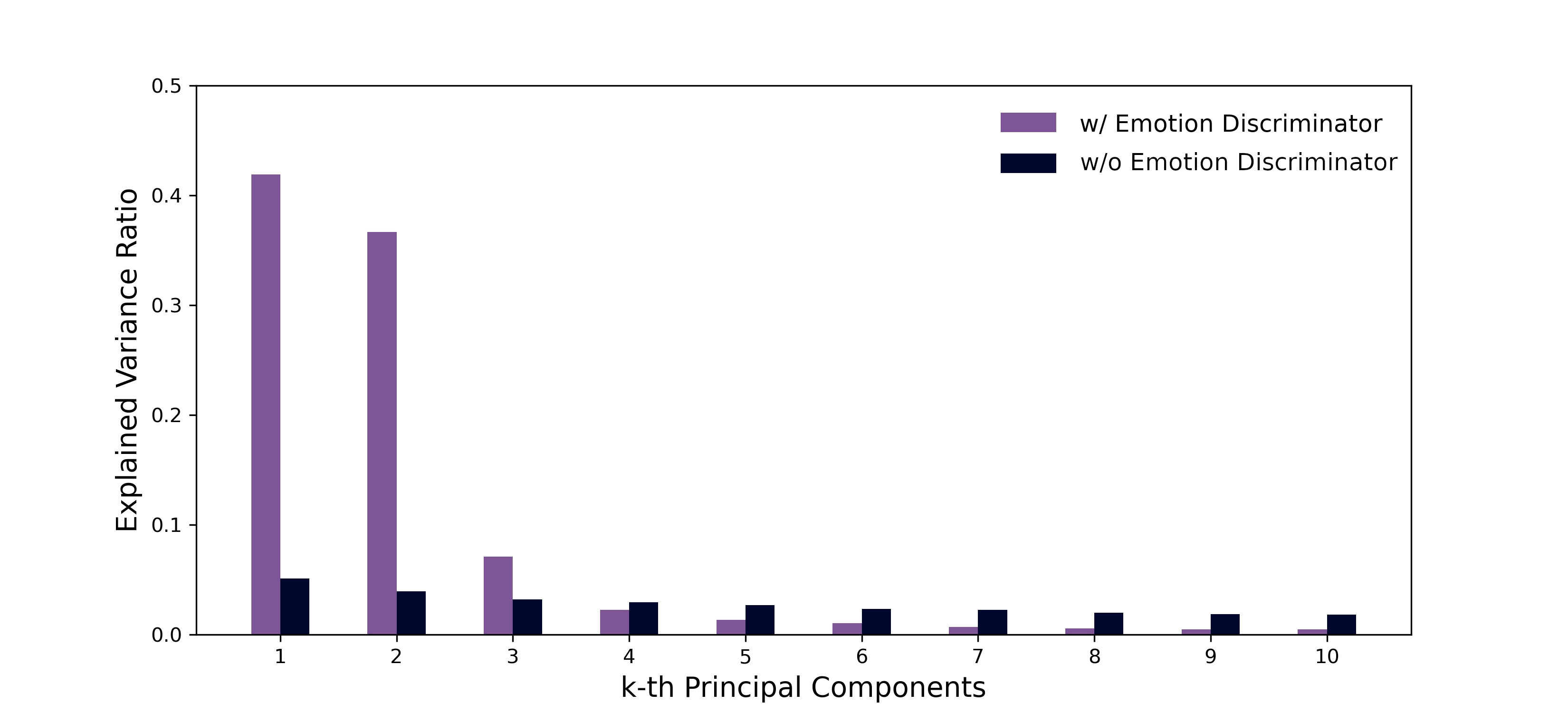}
    \caption{Explained variance ratio of the prosody embeddings from the ablation. Uniform variance in the absence of an emotion discriminator might disallows meaningful decomposition for emotion simulation.}
    \label{fig:ablation-vars-compare}
\end{figure}

Furthermore, if we examine the explained variance ratio across prosody prototypes (eigenvectors), as shown in Figure~\ref{fig:ablation-vars-compare}, it seems like each eigenvector in model not trained with emotion discriminator does not describe as meaningful features as the ones trained with emotion discriminator. Thus, disallowing emotion simulation through prosody embedding decomposition described in Section~\ref{sec:emo-simulate}.

\subsection{Unseen Emotion Transfer}

As we represented emotion with prosody embedding, we found that we can actually simulate unseen emotions to some extent, given arbitrary speech samples. Let $s^{t}$ be the unseen speech sample conveying emotion $\varepsilon_{t}$, we can sample $\textbf{w}^{\varepsilon_{t}}$ by computing the principal components of $s^{t}$, and synthesize the corresponding $u(\textbf{w}^{\varepsilon_{t}})$. This way, not only the synthesized mimic the emotion and prosody, we could also sample a variety of emotion similar to the given sample. We will leave this exploration for future work, but readers can listen to our experimental samples on this idea in our project page.

\section{Conclusion}

In this paper, we introduce Daisy-TTS, an emotional text-to-speech design to model wider spectrum of emotions. Our core method lies in learning and decomposing emotionally-separable prosody embeddings as proxy for emotions. We show that by decomposing and manipulating the learned embeddings, Daisy-TTS capable to simulate emotional speech that conveys: (1) primary emotions, (2) secondary emotions, (3) intensity-level, and (4) polarity of emotions. Through a series of perceptual evaluation, our proposed design demonstrated higher emotional speech naturalness and emotion perceivability compared to the baseline.

\section{Ethics Statement}
The design of Daisy-TTS could be used as a framework to build a highly realistic emotional text-to-speech and voice cloning systems, given sufficient training dataset and appropriate modifications. There are risks for misusing the proposed design for non-consensual and malicious deepfake media generation, specifically ones with the intention of mimicking natural human emotional expression and response.

\section{Limitation} 

Though Daisy-TTS has shown promising results simulating wider spectrum of emotions, there are still notable limitations in this research. Due to the scarcity of emotional TTS dataset, we haven't yet tested whether the proposed framework would perform similarly to other languages. As our emotion representation heavily conforms to the structural model of emotions that requires speech samples expressing a set primary emotions, our method might not be suitable for speech samples that assumes other types of emotion model.





\bibliography{acl2023}
\bibliographystyle{acl_natbib}
\clearpage

\appendix

\section{Appendix}

\subsection{Evaluation Results (Cont'd)} \label{sec:app-eval-result}

\begin{table}[!htb]
    \small\centering
    \begin{tabular}{@{}l@{ ~ }cc@{ ~ }c}
\toprule
 &\multicolumn{3}{c}{Speech Naturalness (MOS; 95\% CI) $\uparrow$}\\
 \midrule
         & \color{violet}\textit{Bittersweet}& \color{violet}\textit{Disappointment}& \color{violet}\textit{Outrage}\\
\midrule
         Ground Truth&  -& -& -\\
         Baseline&   3.433$_{\pm 0.177}$& 3.256$_{\pm 0.221}$& 3.189$_{\pm 0.234}$\\
 Daisy-TTS& \textbf{3.667}$_{\pm 0.183}$& \textbf{3.911}$_{\pm 0.168}$&  \textbf{3.844}$_{\pm 0.185}$\\
 \midrule
 &\multicolumn{3}{c}{Emotion Perceivability (Acc.) $\uparrow$}\\
 \midrule
 & \color{violet}\textit{Bittersweet}& \color{violet}\textit{Disappointment} & \color{violet}\textit{Outrage}\\
 \midrule
 Ground Truth& -& -& -\\
 Baseline& 0.211& \textbf{0.177}& 0.188\\
 Daisy-TTS& \textbf{0.288}& 0.100& \textbf{0.233}\\
 \toprule
    \end{tabular}
    \caption{Results of MOS and emotion perception tests on {\color{teal}primary} and {\color{violet}secondary} emotions, continued from Table~\ref{tab:primary-secondary-emo-mos}.}
\end{table}

\subsection{Secondary Emotions in Structural Model of Emotions}

Through a series of psychological experiments described in~\citep{plutchik1982psychoevolutionary}, the structural model of emotions derived a list of secondary emotions as mixture of primary emotions. Table~\ref{tab:secondary-list} shows the secondary emotions and their corresponding primary emotion pairs. We incorporate this secondary emotions formulation into our proposed model, Daisy-TTS, and experiments.  

\begin{table}[!htb]
    \centering\small
    \begin{tabular}{l|l}
    \toprule
         Primary Emotions&  Secondary Emotion\\
    \midrule
         Joy + Sadness & \textit{Bittersweetness} \\
         Joy + Surprise & \textit{Delight}  \\
         Joy + Anger & \textit{Pride} \\
         Sadness + Surprise & \textit{Disappointment}  \\
         Anger + Sadness & \textit{Envy} \\
         Anger + Surprise & \textit{Outrage} \\


        \midrule
    \end{tabular}
    \caption{Secondary emotions derived from the mixture of primary emotions, as defined in the structural model of emotions.}
    \label{tab:secondary-list}
\end{table}

\subsection{Crowdsourcing Evaluation Setup} \label{sec:mturk-setup}
\begin{figure}[!htb]
    \centering
    \includegraphics[width=1\linewidth]{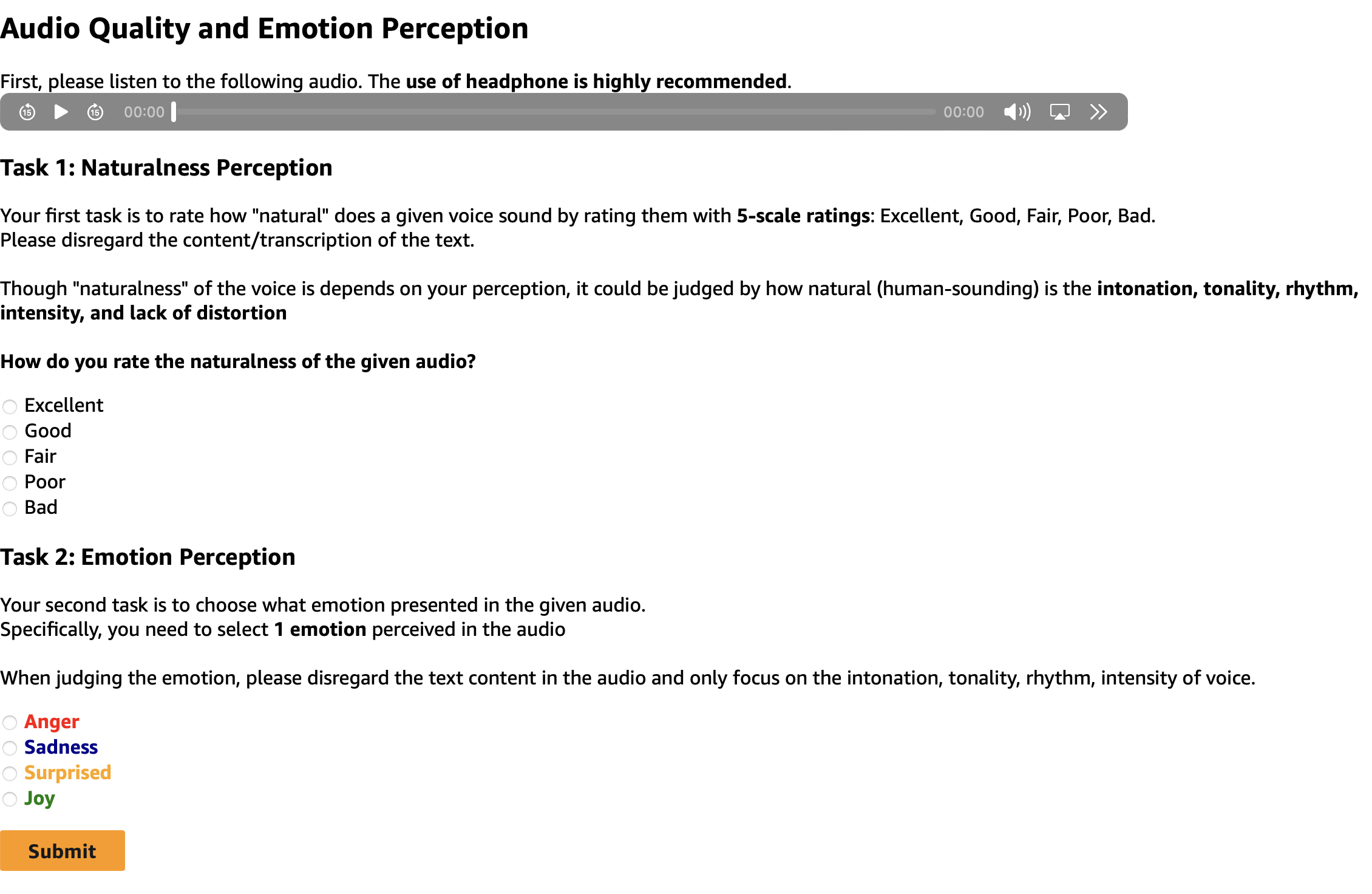}
    \caption{MTurk interface we used for our evaluation.}
    \label{fig:mturk-interface}
\end{figure}

We employ human annotators for our evaluations from Amazon Mechanical Turk. We specify to employ an annotator with HIT of 99\% or above to decrease the chances of unreliable results. The annotators were paid 0.15\$ per instance. The MTurk annotation interface is shown in Figure~\ref{fig:mturk-interface}.


\end{document}